%% file: Research_report.tex
\documentclass[9pt,twocolumn,twoside]{pnas-new}

\templatetype{pnasresearcharticle}
\usepackage{float}
\usepackage{mathtools}

\begin{document}
\title{Electrokinetic Propulsion for Electronically Integrated Microscopic Robots}

\author[a,1]{Lucas C. Hanson}
\author[b,1]{William H. Reinhardt}
\author[a]{Scott Shrager}
\author[b]{Tarunyaa Sivakumar}
\author[b,2]{Marc Z. Miskin}

\affil[a]{Department of Physics and Astronomy, University of Pennsylvania, Philadelphia, PA 19104}
\affil[b]{Department of Electrical and Systems Engineering, University of Pennsylvania, Philadelphia, PA 19104}

\significancestatement{To carry out complex tasks or adapt to changing environments, microrobots must be able to use on-board circuits and sensors to control the actuators that move them around.  Unfortunately, the fastest, most robust, and/or easiest to build approaches to locomotion at the microscale offer no clear path to electronics integration, limiting the complexity of the resulting robot.  Here we show how to bridge this gap, turning simple electrokinetic micromotors into agile, easy to build, electronically controlled microrobots.  The resulting robots live longer, move faster, and are easier to build and control than existing electronically integrated designs.  These results help clear the way for sophisticated, nimble autonomous microrobots that operate without human supervision.}


\authorcontributions{L.C.H., W.H.R., and M.Z.M designed the research.  L.C.H. and W.H.R. designed and fabricated the robots.  L.C.H. characterized the propulsion mechanism, with assistance from S.S.  W.H.R. built the SLM system and programmed the control software, with assistance from T.S, and characterized the kinematics for two engine robots.  M.Z.M. developed the fluid model and calculated numerical solutions for angle of attack and gap height, acquired funding, and supervised research.  L.C.H., W.H.R., and M.Z.M. designed figures and wrote the manuscript.}

\authordeclaration{The authors declare no competing interests.}

\equalauthors{\textsuperscript{1}L.C.H. and W.H.R. contributed equally to this work (listed alphabetically).}
\correspondingauthor{\textsuperscript{2}To whom correspondence should be addressed: Marc Z. Miskin. 
\newline
E-mail: mmiskin@seas.upenn.edu
\newline
Telephone no: 2157466850
\newline
Address: 200 South 33rd Street, Philadelphia, PA 19104 }

\leadauthor{Hanson \& Reinhardt}
\keywords{Microrobots $|$ Electrokinetic propulsion $|$ Micromotors}

\begin{abstract}
Semiconductor microelectronics are emerging as a powerful tool for building smart, autonomous robots too small to see with the naked eye.  Yet a number of existing microrobot platforms, despite significant advantages in speed, robustness, power consumption, or ease of fabrication, have no clear path towards electronics integration, limiting their intelligence and sophistication when compared to  electronic cousins.  Here, we show how to upgrade a self-propelled particle into an  an electronically integrated microrobot, reaping the best of both in a single design. Inspired by electrokinetic micromotors, these robots generate electric fields in a surrounding fluid, and by extension propulsive electrokinetic flows.  The underlying physics is captured by a  model in which robot speed is proportional to applied current, making design and control straightforward.  As proof, we build basic robots that  use on-board circuits and a closed-loop optical control scheme to navigate waypoints and move in coordinated swarms at speeds of up to one body length per second.   Broadly, the unification of micromotor  propulsion with on-robot electronics clears the way for robust, fast, easy to manufacture, electronically programmable  microrobots that operate reliably over months to years.
\end{abstract}

\dates{This manuscript was compiled on \today}
\doi{\url{www.pnas.org/cgi/doi/10.1073/pnas.XXXXXXXXXX}}

\maketitle
\thispagestyle{firststyle}
\ifthenelse{\boolean{shortarticle}}{\ifthenelse{\boolean{singlecolumn}}{\abscontentformatted}{\abscontent}}{}

\firstpage[11]{3}


\dropcap{R}emarkable advances in electronics have laid a foundation for intelligent robots too small to see with the naked eye.  In the past decade, circuit designers pushed computing \cite{wu_004mm316nw_2018,blutman_low-power_2017, do_area-efficient_2019}, memory \cite{wu_004mm316nw_2018,blutman_low-power_2017} , and sensing \cite{wu_004mm316nw_2018,lee_250_2018,atzeni_260274_2022,lim_light_2021,neely_recent_2018} systems into sub-mm dimensions, potentially allowing intelligent decision  making to be carried out on a microrobot \cite{xu_210x340x50um_2022}.   Taking advantage of this progress, roboticists have demonstrated microrobots with increasingly impressive capabilities \cite{huang_increasingly_2022} including autonomous, reconfigurable gait patterns \cite{reynolds_microscopic_2022}, machines that perform two-way optical communication \cite{cortese_microscopic_2020}, and energy transfer to tiny robots with both light and radio frequency fields \cite{miskin_electronically_2020,bandari_flexible_2020,reynolds_microscopic_2022,kim_propulsion_2020}.  Miniaturization of both the circuits and actuators has also dramatically reduced operating power: all the components of a 100-um programmable robot capable of sensing and on-board computing can run on solar power harvested from  ordinary daylight\cite{xu_210x340x50um_2022}.  These growing abilities signal an emerging breed of  microrobots able to sense, adapt, and overcome uncertainty without the need for human supervision.

While prior work in electronically integrated microrobots has centered around a limited set of actuation schemes \cite{reynolds_materials_2024}, the enabling circuits have the potential to improve a number of  microrobot designs. Electrical energy can be converted into a variety of different domains, suggesting that  other approaches to actuation, like those based on chemical\cite{feng_advances_2023}, magnetic\cite{zhou_magnetically_2021}, or acoustic fields\cite{xiao_acoustics-actuated_2022}, could be brought within this framework by establishing on-robot electronic control over the underlying propulsion process.  Building these bridges would enable a more modular, systems-oriented design of microrobots, where individual parts for sensing, computation and actuation can be  mixed and matched, provided they meet electrical and manufacturing constraints.  Further, unifying diverse approaches through electronic control would offer new ways to offset weaknesses and emphasize strengths across platforms.  

Here we take a first step down this road, transforming electrokinetic micromotors into an electronically controllable microrobot actuator.  Already an inherently electrical process, electrokinetic propulsion offers unique advantages like a high speed, low power, and easy fabrication \cite{moran_phoretic_2017,fernandez-medina_recent_2020}.  While typically the electric fields that drive motion are produced by specially tuned chemical reactions, we show they can also be driven by on-robot semiconductor microelectronics.  This fusion of  electrokinetics with circuits brings reciprocal benefits.  Compared to existing actuators for electronically integrated microrobots, electrokinetic propulsion offers faster speeds, longer lifetimes, and simplified fabrication.  Conversely, compared to bare-bones, self-propelled particles, electronics make the resulting robot more robust to chemical changes in its environment, easier to control, and clear a path to integrating sophisticated systems for sensing and computation.  

\section*{Results}
\begin{figure}
\centering
\includegraphics[width=.9\linewidth]{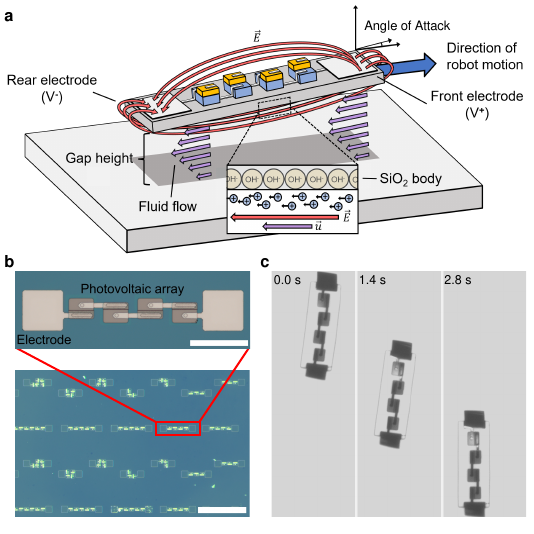}
  \captionsetup{width=.9\linewidth}

\caption{\textbf{Electrokinetic propulsion for microrobots} \textbf{(a)}, Schematic of the electrokinetic mechanism. Positively charged ions in the electrical double layer migrate in the presence of an electric field, $\Vec{E}$, and generate fluid flows around the device that cause locomotion. \textbf{(b)}, Micrograph of a 4 PV design with Ti/Pt electrodes at both ends of the device's SiO$_2$ body (scale bar 100 $\mu$m). Beneath it, a micrograph of various robot designs on chip, fabricated massively in parallel with 400 devices per 1.5 cm square chip (scale bar 500 $\mu$m). \textbf{(c)}, Timelapse of a device locomoting under global microscope illumination in 5 mM hydrogen peroxide.}
\label{fig1}
\end{figure}

The basic mechanism behind electrokinetic propulsion is well established \cite{lyklema_fundamentals_2000,hunter_foundations_2001}  and depicted in Figure \ref{fig1}a.  The motor generates an electric field by feeding current into solution.   This field pushes on nearby mobile charges (e.g., in the electrical double layer surrounding  the motor and/or nearby surfaces or the diffusion layer).  Movement of the charges is resisted by drag from the surrounding fluid,  establishing a flow that moves the motor.

To realize an electronically controlled version, we build robots  that  generate  propulsive electric fields using on-board photovoltaic cells (PVs) (Fig. \ref{fig1}b).  We deliberately keep the circuit simple, stripping the robot's electronics to the bare minimum to facilitate characterization.  The PVs, wired in series, feed current into the solution through 70 x 70 $\mu m^2$ titanium-platinum electrodes at either end of the robot.  Because the nominal voltage supplied by these cells is large enough to perform hydrolysis, the complete circuit is current limited by the incident light flux, allowing us to directly control the applied electric field with the illumination intensity.  For insulation, the PVs and associated wiring are covered in a layer of  photoresist.  All of these parts are fabricated  massively in parallel (see SI Appendix, Section \ref{fab}) with a  previously developed, fully lithographic protocol \cite{miskin_electronically_2020,reynolds_microscopic_2022}, enabling several hundred devices per 1 cm chip as seen in Figure \ref{fig1}b.

\begin{figure*}
\centering
\includegraphics[width=.8\linewidth]{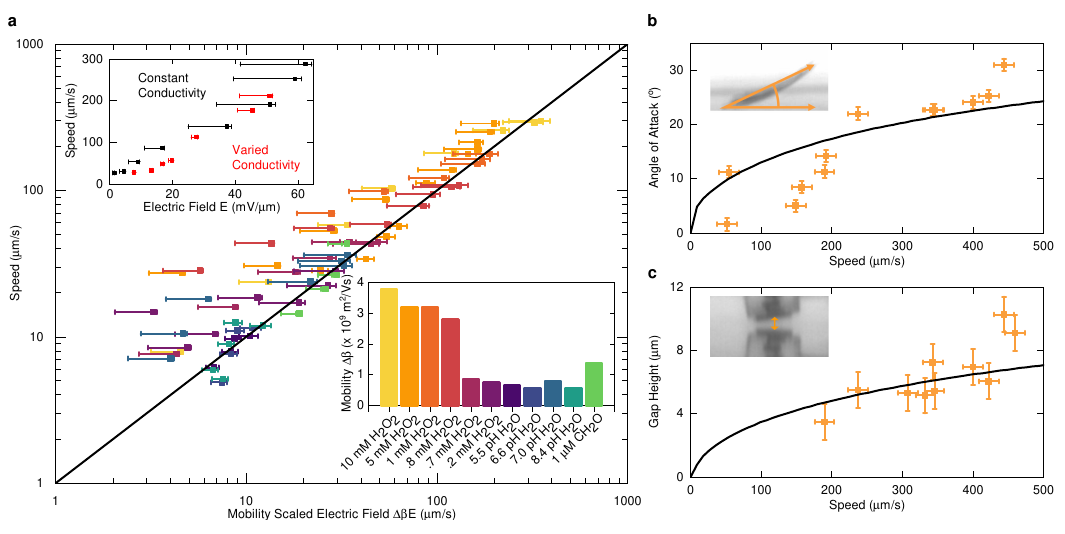}
\caption{\textbf{Characterization of propulsion mechanism} \textbf{(a)}, The upper inset shows the speed vs electric field data for a robot in 5 mM hydrogen peroxide, where the electric field strength is estimated as the quotient of current density and solution conductivity adjusted by a scale factor set by the geometry of the robot (see Methods). The main data are the speed vs electric field behavior for a variety of solutions scaled by the robot's effective mobility $\Delta\beta$, which is determined by linear fitting of the raw speed vs electric field data for each solution (see Methods).  The black line is unity.  The lower inset details the variation of the effective mobility for different solution compositions.  The electric field error bars are the propagated measurement uncertainties of current and conductivity, where the negative skew in these errors is a due to the larger uncertainty associated with solution contamination, which can only cause conductivity to increase.   \textbf{(b)},\textbf{(c)}, Data for the angle of attack and gap height vs speed, respectively, for a robot in 5 mM hydrogen peroxide.  The black curves are the results of numerical fitting of the data to our fluid model (see SI Appendix, Section \ref{propmodel}).  The insets show micrographs of a moving robot viewed from the side and front, from which we measured the angle and gap height.  The error bars are uncertainty estimates set by the resolution of the cameras (see Methods).}
\label{fig2}
\end{figure*}

Once released into solution and illuminated, robots move  at a steady speed, oriented with the positive electrode in the front (Fig. \ref{fig1}c and SI Appendix, Movie S\ref{singleengine}).  Note, as they are negatively buoyant, robots move along the bottom surface of the container they are in.  We independently measure the current produced by the PVs  under various illumination conditions (see Methods) and find  the robot's speed is proportional to current density and inversely proportional to solution conductivity, $\sigma$ (SI Appendix, Fig. \ref{speedvsjsigma}).  Indeed when speed is plotted against the electric field (i.e. the ratio of current density to conductivity),  the data collapse to a single line, as shown in the upper inset of Figure \ref{fig2}a.

Aside from conductivity, we find propulsion speed is largely independent of the surrounding chemical environment.  Figure \ref{fig2}a shows robots can move in a variety of solutions, including two decades of different hydrogen peroxide concentrations, deionized water ($\sigma=$ 250 nS/cm),  salt solutions ($\sigma=$ 10 $\mu$S/cm), pH buffers (5.5 to 8), and formaldehyde.
We also find robots  move on a variety of substrates, including polystyrene, oxygen plasma cleaned glass,  positive charge functionalized glass, SU-8 photoresist films, platinum, and through microfluidic channels (see SI Appendix, Movie S\ref{microchannel}).  In every case, the relationship between speed and electric field is linear, with the slope  alone coupled to the chemical environment (Fig. \ref{fig2}a).  This effect is also small: the highest effective mobility (10 mM hydrogen peroxide) and the lowest (deionized water) only differ by a factor of 3.  By contrast, the speed of the robots can be changed by more than two orders of magnitude through electronic control of the current, effectively decoupling chemical conditions from the robot's ability to move.

To better analyze the underlying fluid dynamics, we image propulsion from the side, directly measuring the robot's  three degrees of freedom: speed, angle of attack, and gap height above the substrate (see SI Appendix, Movie S\ref{sideview}).  As shown in Figure \ref{fig2}, each parameter increases as the field gets larger.   Given the three kinematic constraints from force and torque equilibrium, a model can be constrained to predict a unique solution for the robot's configuration.

We  find three physical ingredients are sufficient to predict the trends in Figure \ref{fig2}.  First, we make a lubrication approximation,  assuming fluid forces are largest in the small gap between the robot and substrate.  All other forces from the fluid are lumped into a small phenomenological forcing term derived by symmetry considerations.  Second, we assume the electric field dominantly points along the axis running from the positive to negative electrode.  Third, we use the ``standard model'' of electrokinetics \cite{lyklema_fundamentals_2000,hunter_foundations_2001}, which couples the fluid to the electric field by imposing a boundary slip condition $\Delta U = \beta \vec{E}$, where $\beta$ is the slip coefficient for that surface.  Note, $\beta$ is the only parameter in this model that depends on the chemical environment, and relates directly to the measured mobility.  In other words, the model behavior is agnostic to the specific ion type involved in the underlying electronic transport through the fluid, a consequence of the fact that the fields are  controlled directly by the robot's circuitry.  We numerically solve the governing fluid and electrical field equations in the gap under the robot (see SI Appendix, Section \ref{propmodel}) and impose force and torque balance to identify equilibrium  angle and gap height at a given electric field strength.

Fitting the model produces the black traces in Figure \ref{fig2}b and Figure \ref{fig2}c, which agree with the data.  Further, the fit parameters are consistent with the underlying physics.  For instance, the difference in mobility between the robot's body and the substrate is a free parameter in the model and the best fit result is within the range of literature values for the mobility of silicon dioxide and polystyrene in aqueous solutions at similar pH values \cite{tuin_electrophoretic_1996,ermakova_effect_2003}.  We note this model is distinct from those used to describe micromotors because we are operating at a much larger length scale: whereas micromotors can have features on the order of the Debye length (here $\sim10-100$ nm), our robots are over three orders of magnitude larger in size and thus in the ``thin Debye layer'' limit.  Further discussion is included in the SI appendix. 

Compared to generic micromotors, those integrated with electronics see several  improvements.   Many motors such as bimetallic nanorods or Janus particles, while elegant in their simplicity, operate at small sizes, low energy scales \cite{paxton_chemical_2006,wang_understanding_2013,paxton_motility_2005}, and are strongly tied to environmental chemistry for propulsion.  By contrast, electronic control of the current flowing through solution lifts these constraints, enabling operation in a range of environments and at larger sizes and energy scales.  For example, the robots presented here operate at fields nearly two orders of magnitude larger than chemically driven motors \cite{paxton_motility_2005}.  By extension, the robot retains the ability to propel at roughly one body length per second, despite the fact that it is more than one hundred fold larger in size.  Likewise, electronically controlled versions can operate in one hundred fold higher conductivity solutions for a fixed robot size.  

These motor performance advantages are further compounded by the fact that electronics offer a  path to complex systems for control and alternate mechanisms of power transfer.  For instance, current controllers can be built a variety of ways, allowing a robot to directly modulate its propulsion speed on command or in response to sensory data.  Further, instead of PVs, acoustic or magnetic fields could be used to transfer energy to on chip electronics \cite{singer_wireless_2021} and drive the electrokinetic propulsion system.  The only fundamental design constraint is to match the voltage and current requirements listed here.  

\begin{figure*}[t]
\centering
\includegraphics[width=.8\linewidth]{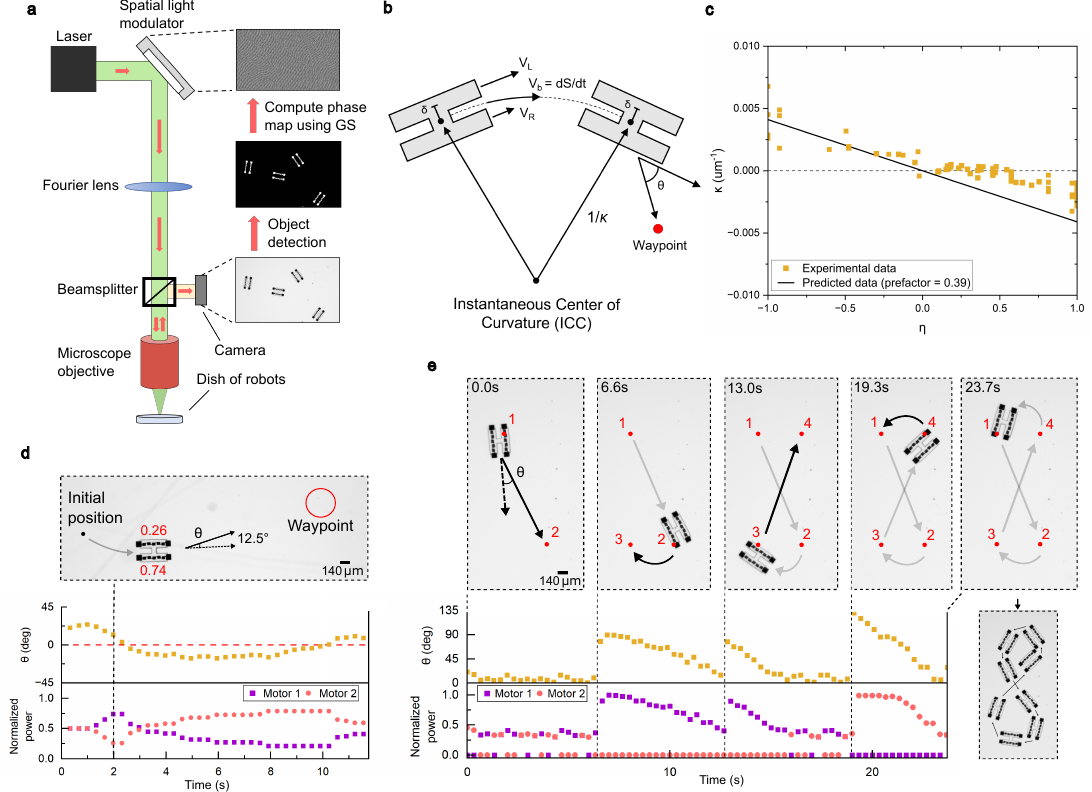}
\caption{\textbf{Optical control and kinematics of a two engine robot} \textbf{(a)}, The closed-loop optical system that allows for automated control over robots in the microscope FOV. After each frame capture, the computer performs object detection, generates a new optical pattern using the Gerchberg-Saxton (GS) algorithm, and updates the spatial light modulator to shoot light at the newly detected robot positions. \textbf{(b)}, A two-motor robot traveling to a waypoint has a body velocity of $V_b$ with engine separation 2$\delta$. Robots travel along arcs of curvature $\kappa$. The misalignment angle $\theta$ is the angle between the robot’s heading and the vector that points from the robot’s center of mass to the target position. \textbf{(c)}, Values of curvature $\kappa$ extracted from multiple experiments show the robot's turning behavior linearly depends on the normalized difference between engine velocities $\eta$. The black line gives predicted values of curvature for a differential drive robot with an additional prefactor of 0.39 to account for effects from stray light (see SI Appendix, Section \ref{kapmodel}). \textbf{(d)}, Implementation of a standard differential drive controller. Both motors are powered simultaneously and the power in each is trimmed in order to obtain propulsion towards the waypoint with a low misalignment angle ($\theta < 15^\circ$). See SI Appendix, Movie S\ref{singleturn} \textbf{(e)}, Path trace of a proportional controller capable of directing the robot to a series of waypoints by powering the left or right motor with a value weighted by the misalignment angle $\theta$. If the misalignment angle is less than a user-defined threshold (here 7.5$^\circ$) for consecutive frames, both motors get the same power simultaneously. See SI Appendix, Movie S\ref{figure8}}
\label{optics}
\end{figure*}

To demonstrate the potential of using circuits to directly control and modulate electrokinetic propulsion in response to real-time data, we demonstrate feedback-controlled steering of individual microrobots.  While the single motor depicted in Figure \ref{fig1}c swims straight, robots require controllable motion.  To achieve this goal, we exploit the linear relationship between speed and current, join  two motors together, and independently vary the current supplied to each.   For simple test circuits built from PVs, we do this using a spatial light modulator (SLM) and a closed-loop control scheme, as shown in Figure \ref{optics}a. The SLM generates optical patterns of high and low intensity light in the microscope field of view to power individual motors (see Methods) and a computer generates new optical patterns based on the image data and user prescribed control laws. This system runs autonomously, tracking robots and shooting light at them as they move, providing a rudimentary way to direct power to each robot's sub-circuits.


By studying the robot's behavior under different proportions of current, we find the motion of two joined engines is governed by differential drive kinematics. Specifically, the left and right side of a robot move at different velocities ($V_L$ and $V_R$, respectively), resulting in a forward velocity equal to their average and a curvature proportional to their normalized difference $\eta=(V_L-V_R)/(V_L+V_R)$ (Fig. \ref{optics}b) \cite{dudek_computational_2010}.   By directly controlling the current in each engine and by extension its projected speed, we can compare the predictions for  differential drive kinematics to the measured data as shown in Figure \ref{optics}c. Accounting for imperfect focusing of the light pattern (see SI Appendix, Section \ref{kapmodel}), we find agreement between the predicted kinematics and experimental data without fit parameters. 


Since differential drive systems can be controlled by a variety of well understood laws, this connection allows us to reliably position robots in space and time.  As an example, we implement two control laws \cite{franklin_feedback_2010,lynch_modern_2017} (see SI Appendix, Section \ref{controllaws})  to direct our robots through a series of waypoints. Both controllers proportionally adjust engine power using the angle between the robot’s heading and a vector that points from the robot’s center of mass to the target position, denoted by  $\theta$ in Figure \ref{optics}.  One controller adjusts both motors simultaneously to direct the robot to a target location (Figure \ref{optics}d and SI Appendix, Movie S\ref{singleturn}) and the other adjusts the power on the motor farther from alignment to steer the robot through a "Figure 8" pattern (Figure \ref{optics}e and SI Appendix, Movie S\ref{figure8}). 

Importantly, the kinematics and control laws can be used regardless of the mechanism that modulates propulsion. Thus, circuits for computation, sensing, and current regulation can implement these controllers entirely onboard, removing the need for the SLM and computer tracking.  Indeed, replacing the optical system would also eliminate imperfections in the current control strategy from stray light, resulting in kinematics that match the expected behavior even more closely.  

\begin{figure*}[t]
\centering
\includegraphics[width=.8\linewidth]{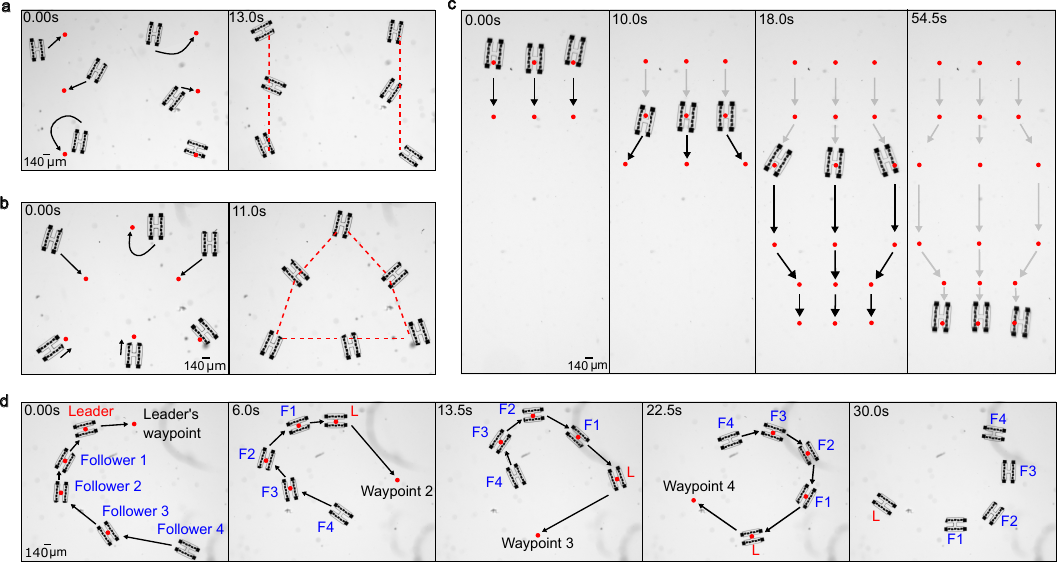}
\caption{\textbf{Addressability and swarming behavior} \textbf{(a)},\textbf{(b)}, The control program can be given a single list of target positions in order to rearrange the robots based on their nearest waypoint. \textbf{(c)}, Robots can be assigned individual lists of waypoints in order to trace out separate paths in unison. \textbf{(d)}, Waypoints can also be dynamic and change at each timestep. Shown in blue, ``follower'' robots are assigned the location of another robot in the system as a waypoint. This rule results in chain-like structures following the ``leader'', shown in red.}
\label{control}
\end{figure*}

Because any circuit, even the simple ones used here, localizes the way energy is converted into motion,  parallel control of many microrobots becomes straightforward. This stands in contrast to current microswimmers, which are inherently difficult to localize and control due to their frequent reliance on bulk energy in the environment for actuation \cite{ebbens_catalytic_2018,zhang_janus_2017}.  To demonstrate, in Figure \ref{control}a and Figure \ref{control}b, we generate a list of waypoints and assign each one to the nearest robot. The robots simultaneously travel to their unique waypoints, rearranging the overall system to form user defined shapes, here lines and triangles (SI Appendix, Movies S\ref{rectrearrange} and S\ref{trirearrange}). Furthermore, each robot can be given its own list of waypoints in order to have multiple robots perform sequences of separate tasks. Figure \ref{control}c and SI Appendix Movie S\ref{blueangels} show three robots swimming in unison on different paths. Last, paths can be dynamic, evolving  with the swarm.  In Figure \ref{control}d,  we create a chain of robots  by assigning each robot to follow a neighboring device  (SI Appendix, Movie S\ref{spiral}).  We note this approach could be readily scaled up either by improving the SLM, or by converting light into a communication channel to reprogram robots dynamically\cite{xu_210x340x50um_2022} (see SI Appendix, Section \ref{scaling}).

\section*{Discussion}

Akin to self-propelled particles, the simplicity of electrokinetic actuators  confers speed and robustness.  We find robots can last over a year in solution, can be  transferred with macroscopic pipettes upwards of one hundred times, and can be dried and re-immersed without loss of function.  This stability marks a  major improvement over other robots we have built using electrochemical  actuators \cite{miskin_electronically_2020,reynolds_microscopic_2022} or bubble generators, which are far easier to damage and typically last for a few weeks at most. Further, electrokinetic actuation is vastly easier to implement, requiring only a single step of lithographic patterning.  The actuator itself is a chemically stable, but more or less arbitrary metal electrode, requiring few material considerations, no atomically thin material layers, and no 3D self-folding construction \cite{reynolds_microscopic_2022,hu_magnetic_2021,liu_micrometer-sized_2021}. These improvements in stability, fabrication, and performance are important for moving microrobots towards real-world applications.  

Crucially, these advantages are realized without sacrificing the capacity to  integrate  electronics.   While the robots here carry the bare minimum circuitry to move, this choice is by design. Circuits are inherently modular, allowing more advanced semiconductor elements to easily integrate and expand functionality.  For instance, the  control laws that govern differential drive could be placed on-robot using already demonstrated computing systems for micro-robotics \cite{xu_210x340x50um_2022}. Indeed, in our prior work we have demonstrated such an evolution,  building simple circuits and actuators to establish a base propulsion technology\cite{miskin_electronically_2020}, and then expanding to advanced circuits for on-robot control \cite{reynolds_microscopic_2022}.

Future work might expand on this 'best of both worlds' scenario by bringing other propulsion mechanisms under electronic control.  For instance, AC electrokinetic techniques are a natural extension of the DC mechanism used here.  Alternatively, it may be feasible to transform resonance and impedance modulation techniques for magnetic  or acoustic power transfer \cite{singer_wireless_2021} into circuit controllable transducers for microrobot  locomotion.  Progress along this front would enable advances in circuits to broadly improve the intelligence and complexity of microrobots, irrespective of how they move.

\section*{Methods}
\input{methods-pnas}

\subsection*{References}
\bibliography{references}
\acknow{The authors would like to thank Michael Reynolds, Brian Leahy, and Matthew Campbell for helpful discussions, and the Singh Center for Nanotechnology staff for technical support during the fabrication process. The authors would also like to thank Professor Tom Mallouk, Professor Paulo Arratia, and Dr. Albane Th\'ery for their comments on the manuscript.  This work was supported by the Army Research Office (ARO YIP W911NF-17-S-0002) and the Air Force Office of Scientific Research (AFOSR FA9550-21-1-0313), and was carried out at the Singh Center for Nanotechnology, which is supported by the NSF National Nanotechnology Coordinated Infrastructure Program under grant NNCI-2025608.}

\showacknow{} 

\clearpage


\section{Supplementary Data}\label{extended}
\subsection{Full Fabrication Protocol}\label{fab}
For our fabrication process (see Supplemental Fig. \ref{fabschem}), we use silicon-on-insulator wafers from Ultrasil corporation with a 2 $\mu$m device layer (p-type doped, 0.1 Ohm-cm), 500 nm silicon dioxide layer, and 500 $\mu$m silicon handle thickness.  15 mm chips are diced from a 4 inch wafer using a dicing saw, spin coated with P509 spin-on glass from Filmtronics, and postbaked for 20 minutes at 120$^{\circ}$C. We use a rapid thermal annealer to diffuse the P509 (n-type phosphorus) dopants by annealing for 30 minutes at 800$^{\circ}$C in an N$_2$ environment to form a layer of n-type silicon above the p-type silicon. Once the doping is complete, we remove the spin-on glass by immersing the chip in 6:1 buffered oxide etch (BOE).

We build the photovoltaics (PVs) through a series of etching and metallization steps. Before starting, we measure the thickness of the silicon device layer precisely at four different locations across the sample using a Filmetrics reflectometer. We spin on Shipley 1813 photoresist and expose using a mask aligner. Next, we develop the exposed photoresist in AZ300 MIF developer for 50 seconds and O$_2$ plasma clean the sample at 150 W for 90 seconds. We etch in an RF plasma SF$_6$/O$_2$ environment for 40 seconds to expose the p-type device layer silicon and then measure the thickness of the silicon again to confirm an etch depth of at least 1 $\mu$m. We strip the photoresist by ultrasonicating in heated PG remover for 10 minutes and O$_2$ plasma cleaning at 150 W for 90 seconds.

Our second etch step forms discrete PVs. First, we spin on an HMDS/AZ3330 photoresist stack and expose using a mask aligner. Next, we develop the exposed photoresist in AZ300 MIF developer for 50 seconds and O$_2$ plasma clean at 150 W for 90 seconds. We adhere our samples to a carrier wafer using CrystalBond and perform an anisotropic Bosch etch with a deep reactive ion etcher. Using the insulating oxide layer of the SOI as an etch stop, we completely clear away the device layer silicon in the exposed regions to form discrete PV cells. We remove the samples from the carrier wafer by ultrasonicating in heated (70$^{\circ}$F) water and then strip the photoresist by ultrasonicating in heated PG remover for 10 minutes. We then O$_2$ plasma clean the samples at 150 W for 90 seconds.

Next, we make metal contacts on the p-type and n-type silicon. We start by insulating the PVs by depositing a 20 nm conformal layer of silicon dioxide using an atomic layer deposition (ALD) tool. We spin on HMDS/AZ3330 photoresist, expose in a mask aligner, and develop in AZ300 MIF developer for 50 seconds to form small openings in the photoresist to the p-type and n-type silicon. We O$_2$ plasma clean at 150 W for 90 seconds then immerse the samples in 6:1 BOE for 15 seconds to remove the insulating oxide layer in the exposed regions. Before the native oxide layer regrows on the exposed silicon, we sputter 20 nm of titanium and 40 nm of platinum over the entire sample.  We then perform a lift-off by stripping the AZ3330 photoresist in heated PG remover with ultrasonication, followed by an O$_2$ plasma clean at 150 W for 90 seconds.  Finally, we anneal the contacts in a rapid thermal annealer for 5 minutes at 400${^\circ}$C in an N$_2$ environment.

To wire the PVs together and form the device's electrodes, we spin on HMDS/AZ3330 photoresist, expose in a mask aligner, develop in AZ300 MIF developer for 50 seconds, and O$_2$ plasma clean at 150 W for 90 seconds. We sputter 20 nm of titanium followed by 40 nm of platinum and perform a liftoff process in heated PG remover with ultrasonication to form interconnects between the PVs and electrodes at either end of the PV line.

Next, we define the shape of the robot's body by etching the silicon dioxide layer of the SOI in the regions between robots. We spin on HMDS/AZ3330, expose in a mask aligner, and develop in AZ300 MIF for 50 seconds. We O$_2$ plasma clean at 150W for 90 seconds and perform a 25 minute CF$_4$ etch to remove the exposed silicon dioxide. We then verify that the oxide has been removed using a Filmetrics reflectometer and strip the photoresist in heated PG remover with ultrasonication.

In order to insulate our electronics stack from the surrounding solution, we spin coat SU-8 2005 epoxy at 3000 RPM for 30 seconds with a 1000 RPM/s ramp rate. We soft bake the samples at 65${^\circ}$C for 1 minute and then at 90${^\circ}$C for 2 minutes before stripping the edge bead and exposing in a mask aligner through an I-line filter. After the sample is exposed, we bake the chips at 65${^\circ}$C for 1 minute, transfer to a separate hotplate to bake at 95${^\circ}$C for 3 minutes, develop for 60 seconds in SU-8 developer with agitation, followed by a rinse in IPA. We then hard bake the SU-8 at 65${^\circ}$C for 1 minute, transferring the samples to a 95${^\circ}$C hot plate, and ramping up to 150${^\circ}$C where the samples are held for 5 minutes. Subsequently, we turn off the hot plate and allow the sample to cool to room temperature to relieve thermal stress in the epoxy.  We O$_2$ plasma clean at 150 W for 90 seconds.

We then sputter 200 nm of aluminum, followed by 10 nm of alumina deposited via ALD. We spin on an HMDS/AZ3330 stack, expose in a mask aligner, and develop in AZ300 MIF for 120 seconds with constant agitation to form small openings to the alumina/aluminum around the perimeter of the robots. After an O$_2$ plasma clean at 150 W for 90 seconds, we post-bake the samples at 115${^\circ}$C for 2 minutes and place them in aluminum etchant A for at least 20 minutes to clear the alumina/aluminum stack and expose the  handle silicon. We remove the photoresist and run a XeF$_2$ etch to remove the underlying silicon through the holes patterned in the aluminum and suspend the devices in air. Lastly, we etch away the alumina/aluminum stack holding the devices by placing the samples in 10:1 diluted aluminum etchant A overnight. Once the aluminum is removed, we pipette the free floating devices out of the etchant and into deionized water for storage.

\subsection{Estimating field strength and mobility coefficient}\label{mobility}

The magnitude of the field in the solution at the electrode interfaces can be computed simply as $J/\sigma = I/a^2\sigma$, where $I$ is the current generated by the PVs, $a$ is the side length of each electrode, and $\sigma$ is the solution conductivity.  However, since the pressures and shears that drive propulsion occur in the fluid gap underneath the robot, we seek to estimate the field strength due to current flowing in this region.  This can be achieved by first estimating the voltage drop in the solution between the electrodes.  For the simple case of a 2D electrode on an insulating substrate with characteristic size $a$ and current density $J$, the potential at the electrode interface relative to a point at infinity can be found from the Laplace equation to be $\Delta\phi\sim I/a\sigma$.  By the linearity of solutions to the Laplace equation, this result holds for the case where a grounding electrode of the same size lies at some lateral distance $l$ from the first electrode, yielding an expression for the potential drop between them.  We can thus approximate the field magnitude as $E\sim \Delta\phi/l=I/a\sigma l=J a/ \sigma l$.  In other words, the propulsive field strength is equal to the field magnitude at the electrode interface reduced by a factor $a/l$.  It is this propuslive field by which the value of the electrokinetic mobility is determined.

\subsection{Modeling Microrobot Propulsion}\label{propmodel}

From the side imaging of robots, we observe that the fluid gap underneath the robot is significantly smaller than the robot's length, $l$ or width, $w$.  This allows for a considerable simplification of both the electrical and fluid parts of the problem by focusing on the field and flows within the gap, as these should dominate the overall behavior of the system.  

We approximate the gap profile as a linear function  along the direction of transit (here the x-axis) and as a constant along the orthogonal axis (the y-axis, defining the robot's width).  The z-axis points along the gap direction.  A schematic of our coordinate system is included as Extended Data Figure \ref{modelcoords}.

In the electrical domain, we assume electrical current flowing across the robot's body is governed by Ohm's law $\nabla^2 \phi =0$.  Since there is a no flux boundary for current both at the lower substrate and the robot's body, we expect current flow along the z-axis (i.e. along the gap direction) to be negligible. Thus, we integrate the potential from the top to the bottom of the gap, reducing the electric field equation to 
\begin{equation}\label{field}
\partial_x (h(x) \partial_x \phi ) +\partial_y (h(x)\partial_y \phi)=0
\end{equation}

Ignoring field variations across the robot's width (y-axis), as no bias is applied along this direction, we find the field is approximately governed by 
\begin{equation}
\partial_x (h(x)\partial_x \phi )=0
\end{equation}

Enforcing that the potential equals the applied electrode potential at either end of the robot leads to the solution
 \begin{equation}
\phi(x) = (\Delta \phi)[\frac{\log[h(x)]}{\log[h_{+}/h_{-}]}-\frac{(h_++h_-)}{2\alpha} \log[\frac{h_++h_-}{2}]]
\end{equation}
where $h_{+}$ is the largest height of the gap, $h_{-}$ is the lowest and $(\Delta \phi)$ is the applied potential drop.  We note the first term conveys all the relevant physics since it is the only part of the expression with spatial dependence and the field alone contributes to propulsion.  The second term is simply a reference potential chosen so that in the limit $\alpha \rightarrow 0$ the potential is approximately $\phi(x) \approx (\Delta \phi) x/l$.

Turning to the fluid domain, forces on the robot are computed by contracting the fluid stress tensor, $\sigma_{ij}=-p(x)\delta_{ij} + \mu (\partial_i u_j +\partial_j u_i)$, with the normal vector on the robot's body, $\hat{n} =\alpha /\sqrt{1+\alpha^2}\hat{x}-1/\sqrt{1+\alpha^2}\hat{z}$.  Here $p$ is the pressure, $u_i$ is the velocity of the fluid flow in the $i$th direction, and $\mu$ is the fluid viscosity.  In the lubrication approximation, we assume that $p$ is independent of z, that the dominate viscous stress term is $\partial_z u_x$, and that the fluid velocity along the x direction is given by $u_x = \frac{1}{2\mu}\partial_x p (z^2-zh(x)) +(V+\Delta \beta E_x)z/h +\beta_{-} E_x  $, where $\Delta \beta$ is the difference in surface slip coefficients, $\beta_{-}$ is the slip coefficient of the surface under the robot, and $E_x=-\partial_x \phi(x)$.  The resulting lift, drag and torque equations respectively read \cite{leal_advanced_2007, szeri_fluid_2011, szeri_pivoted_1970} 

 \begin{align}\label{eos}
f_z = \langle p \rangle +\mu \alpha \langle (V+\Delta \beta E_x) /h(x) \rangle \\
f_{x} = \frac{\alpha}{2}  \langle p \rangle +\mu \langle (V+\Delta \beta E_x) /h(x) \rangle \\
\tau= \langle x p \rangle +\mu \alpha \langle (V+\Delta \beta E_x) x/h(x)\rangle
\end{align}

where angle brackets denote integration over the robot's area $\langle p \rangle =\int_{-l/2}^{l/2}dx \int_{-w/2}^{w/2} dy p$.    The right hand sides represent forces from the lubrication zone, whereas the left hand side represents any external forcing.  

Solving the fluid forces requires computing the pressure under the robot.  Within the lubrication approximation, the pressure satisfies  the Reynolds equation \cite{leal_advanced_2007, szeri_fluid_2011, szeri_pivoted_1970}: 

\begin{align}\label{reynoldsGen}
\frac{1}{12\mu}\partial_i ( h^3\partial_i p)=\frac{1}{2}\partial_i (h (u_i^+ +u_i^-))+u_z^+-u_z^{-}-u^+_i\partial_i h
\end{align}
where $u_i^+$ represents a fluid velocity in the $i$th direction on the upper surface,  $u_i^-$ denotes fluid velocities on the lower surface and contracted indices sum only on the x and y directions (i.e. the pressure is  a function only of x and y).

To simplify the calculation, we exploit the linearity of low Reynolds number flow and split the pressure into  two contributions: one from the translations and rotations of the robot's body ($p_m$) and the other from the electric field induced boundary slips ($p_{ek}$).  These  different boundary conditions for the fluid each set different terms on the right hand side of equation \ref{reynoldsGen}.  

For the flow from rotation and translation, the boundary conditions simplify considerably.  Here all lower surface velocities $u_i^-=0$ while on the upper surface, there is motion from rigid translation in the x and z directions and a spatially varying velocity due to rotation around the center of mass.  Specifically, $u_x^+=V$ and $u_z^+=U+x\dot{\alpha}/(1+\alpha^2))$.  Inserting these expressions gives the equation for the rotation/translation part of the pressure \cite{leal_advanced_2007, szeri_fluid_2011} 

 \begin{align}\label{mechP}
\frac{1}{12\mu}\partial_i (h^3 \partial_i p_m) +  V\alpha/2 = (U+x\dot{\alpha}/(1+\alpha^2) )
\end{align}
 As a boundary condition, pressure is assumed to vanish anywhere outside the surface of the robot, as there is no surface here to support stresses.

For the electrokinetic contribution to the pressure, all the fluid velocities at interfaces are proportional to the electrokinetic slip.  Thus $u_i^+=\beta^+ E_i$ and $u_i^-=\beta^- E_i$.  The resulting Reynolds equation for this part of the flow is:
\begin{equation}
\begin{split}
    \frac{1}{12\mu}\partial_i (h^3 \partial_i p_{ek})  =  \partial_i (h (\beta^{+}+\beta^{-}) E_i)/2 + \beta^+ E_z(z=h) \\ -\beta^ + \alpha E_x(z=h)
\end{split}
\end{equation}

We find all the terms on the right hand side cancel identically.  First, we note that the last two terms on the right hand side cancel, since  the no-flux condition on the electric field through the robot's body implies $\hat{n} \cdot \vec{E}\propto E_x \alpha - E_z =0$.  Likewise, the first term on the right hand side is zero via equation \ref{field}.  Without source terms, the boundary condition of vanishing pressure outside the robot's body means the electrokinetic portion of the flow does not contribute to pressure and it sole forcing is through shear stress.

Since we are not interested in the pressure itself, but rather its integral over the robot's surface, we can numerically compute the forces and torques in equation \ref{eos} using the variational approach in \cite{szeri_pivoted_1970, szeri_fluid_2011}.  
Specifically,   equation \ref{mechP} implies the pressure also minimizes the functional

 \begin{align}\label{minFunc}
\langle [\frac{\partial_i p h^3 \partial_i p}{12 \mu} -p (V \alpha /2-U-\dot{\alpha}x/(1+\alpha^2))]\rangle
\end{align}

Thus, we approximate $p\approx (y^2 - (w/2)^2)\sum_{n=1}^N \sin[n\pi (x+l/2)/l]a_n$ where $a_n$ are free parameters choosen to minimize equation \ref{minFunc}.   This expansion takes the first two terms of a Taylor series in the y coordinate, accounting for symmetry and boundary conditions, and expands pressure variation along the x-axis in terms of trigonometric functions that vanish at the ends of the robot. Inserting this form into equation \ref{minFunc} and minimizing reduces finding the pressure to a sparse linear algebra problem, which is efficient to solve numerically.

For the external forcing in equation \ref{eos}, we assume $f_z=mg$ and introduce two phenomenological parameters $\tau = a \Delta \phi$ and $f_x= b \Delta \phi$.  These parameters are motivated by additional experiments in which robots on their sides were observed to translate and rotate in proportion to the incident optical power.  

To find the robot's equilibrium state, we insert the pressure expansion and external force parameters $a$ into equation \ref{eos} with the optimal $a_n$.  The three equations can be used to solve for the velocity in terms of the field and forcing parameters alone.  This result can be reinserted into equation \ref{eos} to give two closed equations for the gap height and angle of attack evolution.  We solve these two equations numerically (using Scipy's ODEInt Package) until they reach equilibrium to find the steady state speed, angle of attack, and velocity.  

To produce the fits in Figure \ref{fig2}b and c, we used the Nelder-Meade simplex algorithm to simultaneously minimize the mean square error between the observed data and numerical solutions.  Data and the model were written in dimensionless form, scaling all lengths by the robot's longest dimension $l$, all velocities by the viscous settling time $\mu l/mg$, and all forces by $mg$.  The resulting system of equations has three free parameters: the scaling between velocity and electric field and the phenomenological torque and shear parameters $a$ and $b$, which the solver adjusts to minimize error.     
All of the code used for solving for the pressure, integrating the ODE and fitting the data is included in the supplementary material.

\subsection{Model for predicting $\kappa$}\label{kapmodel}
In an ideal differential drive system lateral motion is prohibited and the path of the robot can be described by its progression along a tangent circle with curvature $\kappa$.  The robot travels with a body velocity $V_b$ and the motor separation is given by 2$\delta$. We write the following equations for the left and right motor velocities:
\begin{equation}
    V_L = V_b(1+\delta\kappa)
\end{equation}
\begin{equation}
    V_R = V_b(1-\delta\kappa)
\end{equation}
The body velocity $V_b$ can be written as $\frac{V_L}{2} + \frac{V_R}{2}$. We solve for the curvature $\kappa$ by subtracting $V_L$ and $V_R$ using this substitution for $V_b$.
\begin{equation}
    \kappa = \frac{V_L-V_R}{(V_L+V_R)\delta}
\end{equation}
We previously defined $\eta=(V_L-V_R)/(V_L+V_R)$, so we write $\kappa$ as:
\begin{equation}\label{adjust}
    \kappa = \frac{\eta}{\delta}
\end{equation}
While Equation \ref{adjust} describes an ideal differential drive system, we find empirically that the direct relationship between light intensity and motor velocity requires an additional term to capture the effect of optical power incident on one motor spilling over to the other motor due to imperfect hologram resolution. We extract this term experimentally by measuring the speed of a motor when the optical power is incident on the motor and again when it the laser spot is offset by a distance $\delta$. The ratio of these two values gives the adjustment prefactor. Experimentally we find a value of 0.39 for the devices shown in Figure 3 and Figure 4. To find the predicted values of curvature seen in Figure 3c we calculate Equation \ref{adjust} for the entire range of $\eta$ and multiply by this prefactor.

\subsection{Control Laws}\label{controllaws}
We implement two controllers in Figure 3d and Figure 3e. In both demonstrations, the computer calculates a misalignment angle $\theta$ after every frame capture as the angle between the robot’s heading and a vector that points from the robot’s center of mass to the target position. In the first control scheme, if $\theta$ is increasing or larger than a user-defined threshold (here 15$^\circ$), the controller increases the power on the motor farther from the waypoint to realign the robot. If $\theta$ is decreasing, the power proportion is not changed. Using this rule, the controller maintains a misalignment angle of less than 15$^\circ$ over a majority of the path trace (Fig. 3d). With the same definition of $\theta$, the second controller is a proportional controller that directs the robot by powering the left motor, the right motor, or both when $\theta$ is smaller than a user-defined threshold (here 7.5$^{\circ}$) for two consecutive frames. Additionally, this controller uses $\theta$ to weight the motor power. As seen in Figure 3e, sharp increases in $\theta$ after receiving a new waypoint results in the controller sending all available power to a single motor for turning.

\subsection{Scaling up to more robots}\label{scaling}
A common goal of various microswimmer platforms is to build systems consisting of multiple devices that are capable of exhibiting swarm-like behaviors such as coordinated locomotion. In this work, we show that tying the propulsion system to onboard electronics turns robots into discrete devices that can easily operate separately from one another provided the electrical currents can be independently controlled. While here we do this directly using optics, integrating onboard circuits to regulate the motor currents and perform computation is a direct route towards building large swarms of independently controlled robots, each with its own onboard brain, controller, and propulsion system. 
To interact with robots, circuits for optical communication \cite{cortese_microscopic_2020, reynolds_microscopic_2022} 
can enable the optical signal to be repurposed as a channel to interface with the robots individually. Using light, we can beam instructions to any number of devices in unison to send information, reprogram devices, or transfer data. Notably, when using the optical system in this way, the computation requirements do not significantly change when the system size increases as the runtime of the SLM system is dominated by two, scale-independent steps: Extraction of the phase map with the GS algorithm \cite{christopher_benchmarking_2020} 
and updating the SLM display with a new optical pattern (which is set by the switching speed of the SLM). In other words, we can send data or individual instructions to each robot in a swarm simultaneously by multiplexing light without worrying about computation time as the system size scales up.

\clearpage
\setcounter{figure}{0}
\counterwithin{figure}{section}
\renewcommand{\thefigure}{S\arabic{figure}}

\begin{figure}[H]
\centering
\includegraphics[width=6cm]{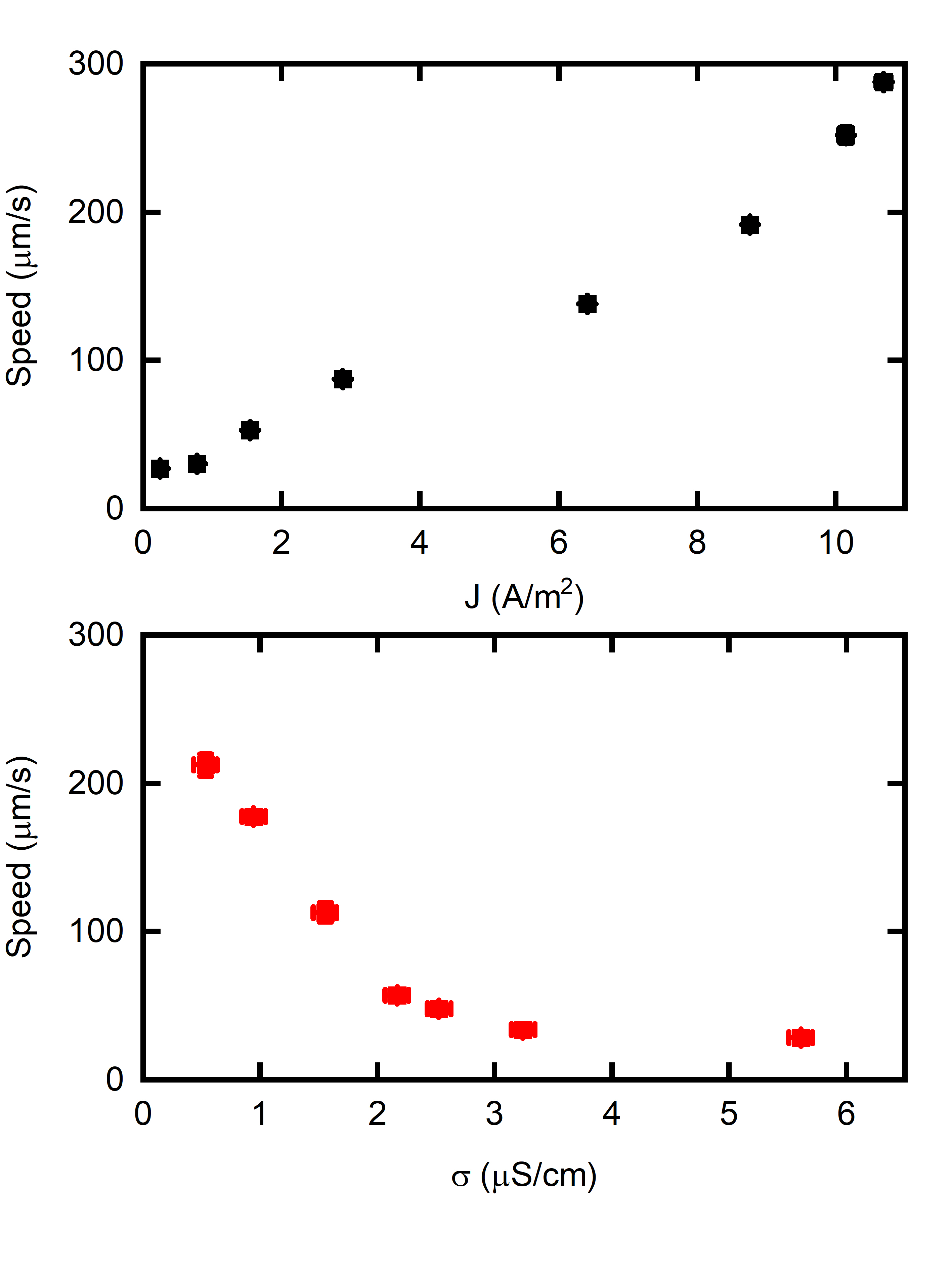}
 \caption{\textbf{Speed vs. current density and conductivity.}
 Top: Robot speed as a function of current density for a fixed solution conductivity of 300 nS/cm in 5 mM hydrogen peroxide.  Bottom: Robot speed as a function of conductivity in 5 mM hydrogen peroxide.  Here current density also varies due to the changes in conductivity.}
\label{speedvsjsigma}
\end{figure}

\counterwithin{figure}{section}
\renewcommand{\thefigure}{S\arabic{figure}}
\begin{figure}[H]
\centering
\includegraphics[width=6cm]{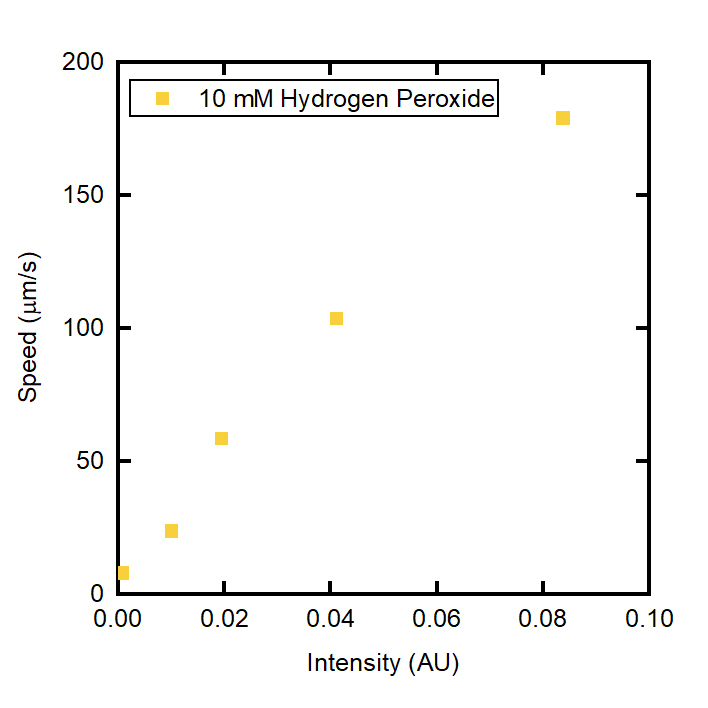}
 \caption{\textbf{Speed vs. intensity.} Robot speed as a function of light intensity, i.e. optical power.  For speeds below approximately 200 $\mu$m/s we find a linear relationship.}
\label{speedvsintensity}
\end{figure}

\begin{figure}[H]
\centering
\includegraphics[width=9cm]{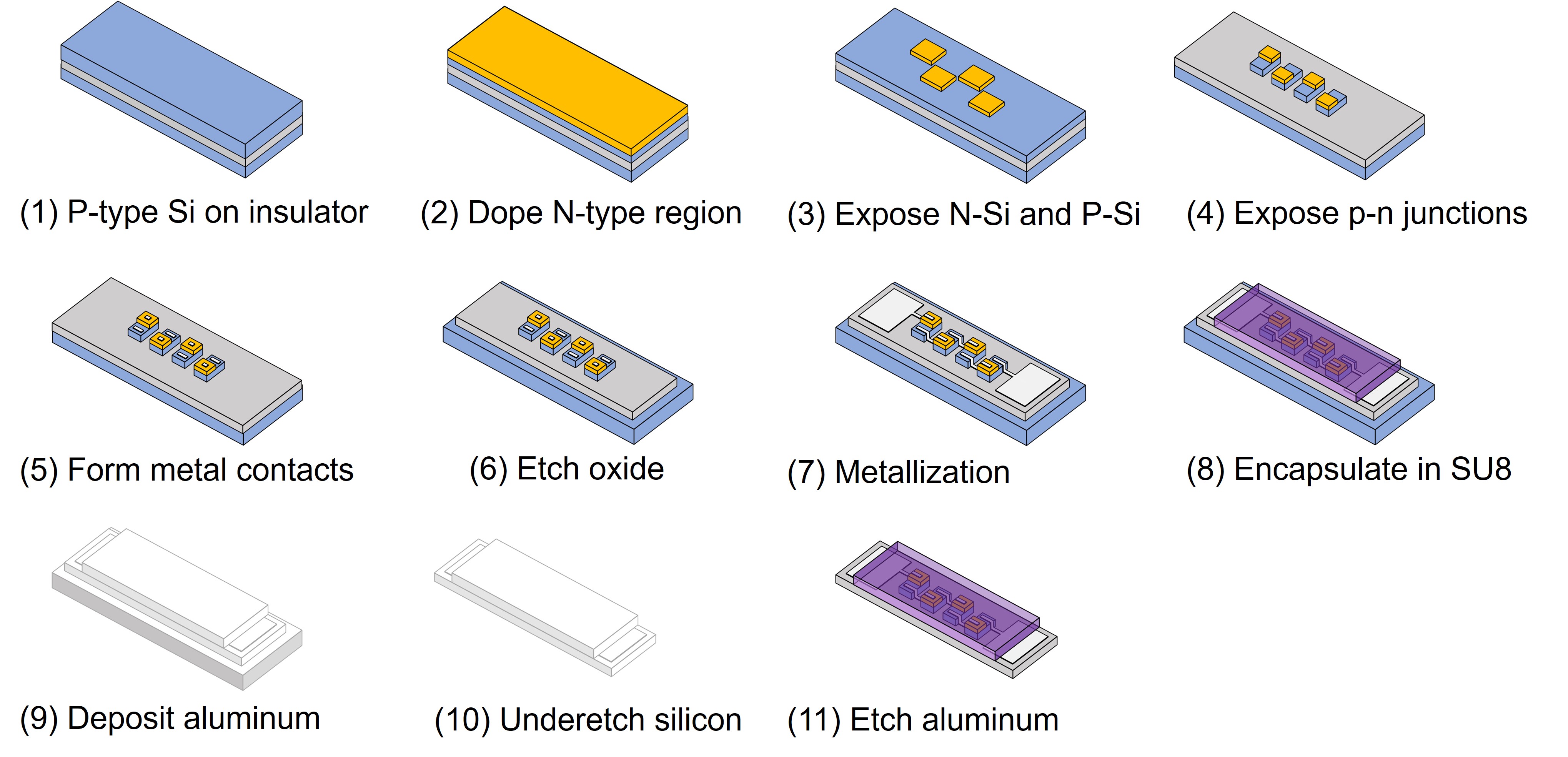}
\caption{\textbf{Fabrication of a single motor.}
Steps (1)-(4) form discrete PVs through a series of doping and etching steps, Steps (5)-(7) deposit the electrical interconnects and actuator electrodes, and Steps (8)-(11) encapsulate and release the motors.}
\label{fabschem}
\end{figure}

\begin{figure}[H]
\centering
\includegraphics[width=6cm]{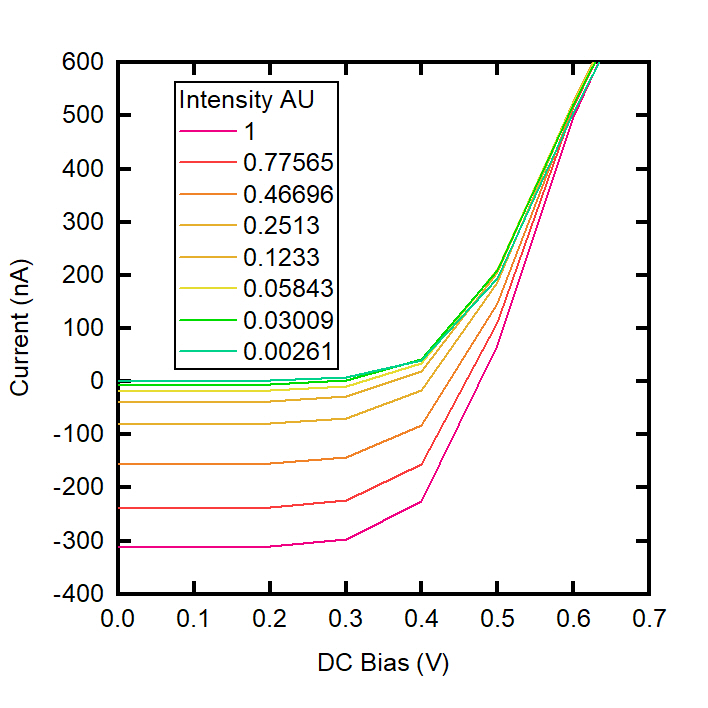}
 \caption{\textbf{PV I-V sweeps. }
 Current vs voltage sweeps of a single PV under various illumination intensities, normalized to the maximum intensity of our light source.}
\label{IV}
\end{figure}

\begin{figure}[H]
\centering
\includegraphics[width=8cm]{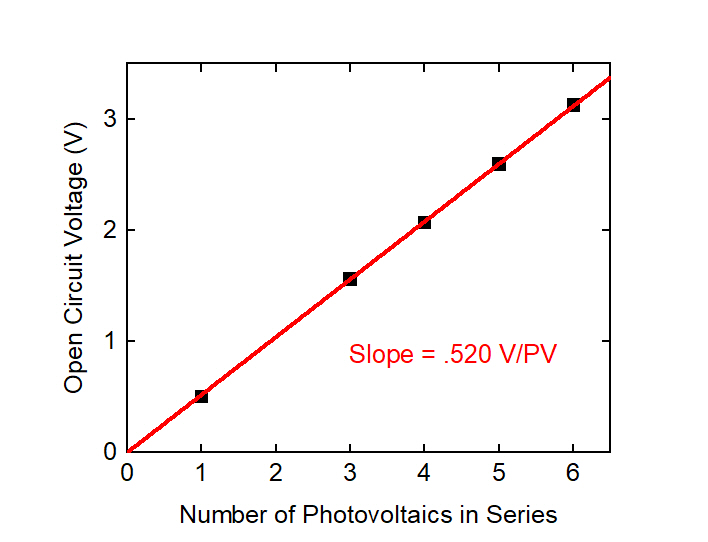}
 \caption{\textbf{Open circuit voltage.} Open circuit voltage at maximum intensity as a function of the number of PVs wired in series.}
\label{OCP}
\end{figure}

\begin{figure}[H]
\centering
\includegraphics[width=6cm]{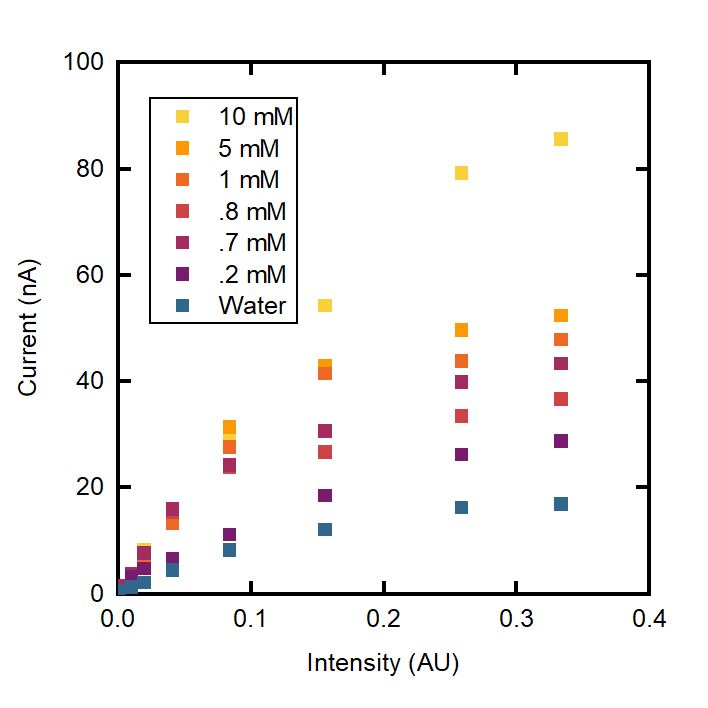}
 \caption{\textbf{PV output in different chemical environments.} Current driven through various solutions by the PVs as a function of light intensity.}
\label{iph}
\end{figure}

\begin{figure}[H]
\centering
\includegraphics[width=6cm]{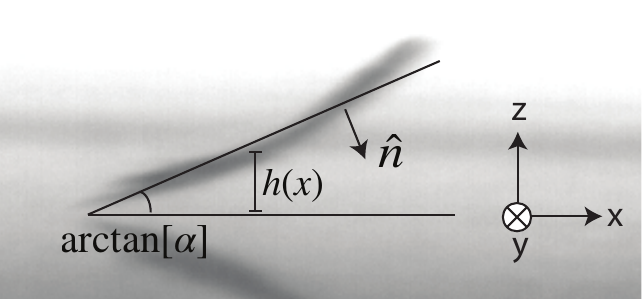}
 \caption{\textbf{Frame of reference for modeling propulsion.} A schematic of the coordinates and approximation for geometry under the robot's body used in modeling the electric field and fluid flows. }
\label{modelcoords}
\end{figure}


\section{\small{Supplementary Video 1}}\label{singleengine}
A single motor under microscope illumination. (This video is in real time).

\section{\small{Supplementary Video 2}}\label{microchannel}
A single motor robot travels through an SU8 channel under microscope illumination. (This video is in real time).

\section{\small{Supplementary Video 3}}\label{sideview}
A side-on view of a two motor robot under illumination. Due to asymmetry in the motors, the robot drives in a circle. At the end of the video, the driving illumination is turned off and the robot stops moving. (This video is in real time).

\section{\small{Supplementary Video 4}}\label{singleturn}
A controller autonomously pilots a robot to a waypoint by trimming the power on each engine. (This video has been sped up by 2x).

\section{\small{Supplementary Video 5}}\label{figure8}
A controller autonomously pilots a robot through a series of waypoints to trace out a Figure 8 pattern. (This video has been sped up by 2x).

\section{\small{Supplementary Video 6}}\label{rectrearrange}
Given a list of target locations, robots autonomously travel to the nearest waypoint to form user-defined shapes. Shown here is a rectangle. (This video has been sped up by 2x).

\section{\small{Supplementary Video 7}}\label{trirearrange}
Given a different list of target locations, robots rearrange to form other user-defined shapes such as triangles. (This video has been sped up by 2x).

\section{\small{Supplementary Video 8}}\label{blueangels}
Robots are assigned an individual list of waypoints to trace out separate paths simultaneously. (This video has been sped up by 2x).

\section{\small{Supplementary Video 9}}\label{spiral}
Robots are assigned dynamic waypoints such as the location of another robot in the system and form a chain-like structure in which they follow their nearest neighbor. Additionally, the controller prevents robots from getting too close together, resulting in the stop-and-go behavior seen in this video. (This video has been sped up by 2x).
\end{document}

%% file: methods-pnas.tex
\subsection{Microrobot fabrication}

Microrobots are fabricated using standard semiconductor processing techniques on p-type silicon on oxide (SOI) wafers, consisting of a 2 micron thick device layer, a 500 nm silicon dioxide layer, and 500 microns of handle silicon.  Photovoltaics are fabricated by diffusing n-type dopants into the the top layer of silicon, and then plasma etching mesa structures to allow the formation of contacts to both the n-type and p-type layer.  Conformal silicon oxide is deposited to insluate photovoltaics, and contacts are formed via HF etching and sputtered titanium and platinum.  Subsequently, interconnects between the photovoltaics, as well as electrodes, are formed from sputtered titanium and platinum.  The photovoltaics and wires are insulated with SU8, and the bodies of the robots are defined by plasma etching the silicon dioxide layer.  The robots are released from the wafer by sputtereing an aluminum support film on top of the robots, and then etching the handle silicon from underneath using XeF2 vapor.  The support film is then etched in aluminum etchant A, resulting in released individual robots.  Our fabrication process is illustrated in SI Appendix Fig. \ref{fabschem}, and discussed in detail in SI Appendix \ref{fab}.

\subsection{Characterizing PVs}

PVs were characterized by probing with tungsten microprobes (Signatone SM-35) and performing current-voltage sweeps at various illumination intensities using a Keithley source meter (2450). Individual PVs output photocurrents on the order of 100 nA and open circuit voltages of approximately 500 mV when illuminated with a white light source (Thorlabs Solis 1-D) at maximum power in an upright microscope (Olympus) (SI Appendix, Fig. \ref{IV}).  Open circuit voltages sum for multiple PVs wired in series, enabling the devices to operate effectively as current sources in solution (SI Appendix, Fig. \ref{OCP}).

\subsection{Solution and substrate preparation}

Hydrogen peroxide solutions were prepared by dilution of 30\% hydrogen peroxide (Fisher H325-500) with DI water from our facility with an initial conductivity of 250 nS/cm.  An upper bound of 50 mM hydrogen peroxide concentration was observed, at which point catalytic bubble formation occurs, disrupting propulsion.  Conductivity of solutions was adjusted with addition of sodium nitrate (Sigma-Aldrich S5506-250G) and sodium nitrite (Fisher S347-500) at various concentrations.  Buffer solutions were prepared by dilution of various buffers (Fisher SB107-500, Fisher SB115-500, Fisher SB101-500), and  50\% sodium hydroxide (Transene 1310-73-2). Formaldehyde solutions were prepared by dilution of 37\% formaldehyde stabilized with methanol (Fisher F79-500). Solution pH was measured with a Hach Pocket Pro pH meter (PN 9531000) and solution conductivity was measured with a Hannah Instruments pure water conductivity meter (HI98197).  Polystyrene substrates used were sterile 60 x 15 mm petri dishes (VWR International 25384-D92).  Glass substrates used were microscope slides (Fisher 12-549-3 and Thorlabs MS10PC1, for negative and positive surface functionalization respectively).  Platinum substrates used were made by sputtering 20 nm of titanium and 40 nm of platinum on a 25 mm glass cover slip (Deckglaser 100).  SU-8 substrates and microfluidic channels used were made by spinning SU-8 2050 photoresist on a 4 inch borofloat wafer at 2000 rpm for 40 seconds, soft baking 9 minutes at 95${^\circ}$C, exposing regions around channels with a mask aligner through an I-line filter, post baking for 7 minutes at 95${^\circ}$C, developing in SU-8 developer for 7 minutes with agitation, rinsing with IPA, and hard baking for 5 minutes at 200${^\circ}$C.

\subsection{Measuring the speed vs. current response of a single motor robot}

To measure the current driven through solution by a robot under various illumination intensities we fabricate two test chips: a test chip with PV circuits identical to those on the robots, and a test chip with identical electrodes.  The electrode test chip is immersed in the desired solution under a stereoscope (ZEISS SteREO Discovery.V8), and the PV test chip is placed under an upright optical microscope in air (Olympus).  We probe the circuits on the PV chip with tungsten probes (Signatone SM-35) under illumination in reflection mode with a variable intensity white LED source (Thorlabs Solis-1D).  We probe the electrodes on the electrode chip with insulated Pt/Ir probes (Microprobes for Life Science PI20031.5A5).  We then wire the probes from each chip in series, such that the PV circuit is driving current through the electrodes in solution, which we measure at various light intensities with a low noise current preamplifier (Stanford Research Systems SR570).  The results of these measurements are detailed in SI Appendix Figure \ref{iph}.  We then place a single motor robot in the same solution used to measure the current, and illuminate the robot at the same light intensities in order to measure the speed.  We take 10 frame per second image sequences of the robot in motion and compute the center of mass of the robot in each frame though image thresholding and particle detection (ImageJ).  The position data is smoothed by convolution with a Gaussian kernel, and the instantaneous speed is calculated for each frame as the magnitude of the position change divided by the time between frames.  The reported speed is the average of all instantaneous speeds for the duration of the experiment, usually a few seconds.  To estimate the mobility coefficient for each solution the speed vs field graphs are fit via linear regression, with the speed at zero field constrained to be zero.  The data in Figure \ref{fig2}a are scaled by the effective mobility to force data collapse.  

\subsection{Measuring angle of attack and gap height}

To measure angle of attack and gap height, a two motor robot is placed on a piece of polystyrene in a glass cuvette (Thorlabs CV10Q35A), and imaged simultaneously from the side using an adjustable magnification microscope (Olympus) in transmission mode, and from above using a stereoscope (ZEISS SteREO Discovery.V8), while illuminated at various intensities with a ring light (Schott S80-55).  The speed of the device is measured using image sequences from the stereoscope, and the angle of attack and gap height are measured using image sequences from the side mounted microscope.

\subsection{Forming computer generated holograms}
To form optical patterns in the microscope field of view, we mimic setups that are capable of reconstructing 2D or 3D holograms by displaying grayscale phase maps on an SLM \cite{kozacki_color_2016, liu_complex_2011, padgett_holographic_2011}. Using a Holoeye LETO-3 SLM in a phase-modulation scheme, we run the Gerchberg-Saxton (GS) algorithm on a grayscale image containing the optical pattern we want to create \cite{madsen_comparison_2022}. This algorithm extracts a phase map that will diffractively reconstruct the target optical pattern in the far-field plane of the microscope after passing through a lens \cite{kim_gerchberg-saxton_2019,zhou_holographic_2023}.
Additionally, to handle swapping between the coordinate system of the microscope FOV (5320 x 3032 pixels) and the SLM display (1920 x 1080 pixels), we generate an affine transformation matrix using an automated calibration program. This program places non-collinear points with known $(x,y)$ coordinates on the SLM, detects the corresponding $(x',y')$ points in the microscope FOV, and generates a transformation matrix to map $(x,y) \leftrightarrow (x',y')$. At each time step, we apply this matrix to the output of the GS algorithm to generate a phase map for the SLM that recreates the desired optical pattern in the microscope FOV with the correct scale, rotation, and shear. Further, as shown in in Figure \ref{optics}a, the SLM is tilted at an angle relative to the incident laser light. This skew steers the 0th order spot, which cannot be modulated by the SLM, off the center of the optical axis. We then apply a blazed grating to the phase map extracted by the GS algorithm to bring the 1st order diffraction pattern back to the center of the optical axis and into the backplane of the microscope \cite{maurer_what_2011}. 

\subsection{Object Detection}
Robot detection is performed in the closed-loop cycle pictured in Figure \ref{control}a. Images from a Basler (Ace2 USB Camera) are sent to a Python script where robot positions and engine locations are determined using adaptive thresholding in OpenCV \cite{bradski_opencv_2000}. Tracking of individual robots frame-to-frame is done using Norfair \cite{alori_tryolabsnorfair_2023}.